# Reinforcement Learning to Discover a North-East Monsoon Index for Rainfall Prediction in Thailand

Kiattikun Chobtham
Hydro-Informatics Institute
Ministry of Higher Education, Science, Research and Innovation,
Bangkok, 10900, Thailand
kiattikun@hii.or.th

*Abstract* - **Accurately predicting long-term rainfall is challenging. Global climate indices, such as the El Niño-Southern Oscillation, are standard input features for machine learning. However, a significant gap persists regarding local-scale indices capable of improving predictive accuracy in specific regions of Thailand. This paper introduces a novel North-East monsoon climate index calculated from sea surface temperature to reflect the climatology of the boreal winter monsoon. To optimise the calculated areas used for this index, a Deep Q-Network reinforcement learning agent explores and selects the most effective rectangles based on their correlation with seasonal rainfall. Rainfall stations were classified into 12 distinct clusters to distinguish rainfall patterns between southern and upper Thailand. Experimental results show that incorporating the optimised index into Long Short-Term Memory models significantly improves long-term monthly rainfall prediction skill in most cluster areas. This approach effectively reduces the Root Mean Square Error for 12-month-ahead forecasts.**

## I. INTRODUCTION

Climate prediction is a challenging task due to complex spatiotemporal relationships within Earth systems. Data-driven machine learning approaches have emerged as promising alternatives, whereas traditional Numerical Weather Prediction (NWP) models are usually computationally expensive. Machine Learning–based Weather Prediction (MLWP) typically uses observed climate variables as input features to predict climate variables at future time steps. These variables are particularly effective for short-term forecasts, ranging from hours to a few days [1]. These tasks are commonly applied in early warning systems for extreme rainfall events.

However, long-term prediction for water management applications such as drought or flood risk, reservoir operations, and agricultural irrigation planning is possible using climate indices as predictors [2]. The El Nino-Southern Oscillation (ENSO) [3] can be represented by the Oceanic Nino Index (ONI). ONI is calculated as the Sea Surface Temperature (SST) anomaly averaged over the Niño 3.4 region (5°S - 5°N, 170°W - 120°W) and is used to identify El Niño, La Niña, or neutral conditions. SST-based indices serve as suitable input features for long-term rainfall prediction rather than winds, because they evolve gradually over time. Other commonly used climate indices include the Pacific Decadal Oscillation (PDO), Indian Ocean Dipole Mode Index (DMI), Boreal Summer Intraseasonal Oscillation (BSISO), Multivariate ENSO Index (MEI), and Madden-Julian Oscillation (MJO) [4, 5, 6]. These climate indices can be considered global-scale or tropical teleconnections. Therefore, to improve the rainfall prediction skill in Thailand, a new climate index needs to be discovered to better reflect regional monsoon circulations.

The monsoon is a local phenomenon that significantly impacts rainfall in tropical Southeast Asia. Previous studies have shown that differences in wind speed between two specific regions (Area A: 10–16.25°N, 110–118.75°E; 8.75–10°N, 108.75–116.25°E; and 7.5–8.75°N, 107.5–115°E in the South China Sea, and Area B: 3.75–6.25°N, 103.75–106.25°E) cause heavy rainfall over the east coast of Peninsular Malaysia during the North-East Monsoon (NEM) period [4, 7]. The NEM has a statistically significant impact on rainfall in Thailand, but the specific NEM areas remain unclear.

In this work, we introduce a monthly NEM climate index derived from SST to provide a better representation of the climatology of the NEM. The calculation regions are optimised through state-action exploration using reinforcement learning. Finally, we demonstrate that the proposed climate index improves long-term rainfall prediction in Thailand when incorporated into state-of-the-art deep learning models.

## II. FORMULATION OF RAINFALL PATTERNS IN THAILAND

Rainfall in Thailand is primarily influenced by two monsoon circulations: the boreal summer monsoon (South-West monsoon) and the boreal winter monsoon (NEM). The timing and intensity of these monsoons vary across regions [7, 8]. Specifically, rainfall in southern Thailand is strongly associated with the NEM circulation, during which northeasterly winds transport cold air from the Siberian High and moisture from the South China Sea and the Gulf of Thailand toward the Malaysia Peninsula from October to March [4]. In contrast, rainfall in upper Thailand is dominated by the South-West monsoon (associated with SST differences between two areas 87°E-91°E, 20°N-21°N and 90°E-95°E, 8°S-10°S) in the tropical Indian Ocean, which typically occurs from May to October, corresponding to the monsoon onset and peak of the rainy season. These two monsoon circulations are often associated with monsoon troughs that cause extreme rainfall.

To analyse the correlations between the proposed NEM index and rainfall for each area across Thailand, observed rainfall from all gauge stations will be grouped into spatial clusters using hierarchical clustering, as described in Algorithm 1, rather than grouping stations by provinces. The algorithm first computes the average monthly rainfall and its locations for each station, followed by normalisation and dimensionality reduction using Principal Component Analysis (PCA). Euclidean distances are then calculated, and clusters are formed by iteratively merging stations whose centroid distances fall below a threshold $d$.

**Algorithm 1:** Hierarchical clustering of rainfall stations

**Input**: monthly rainfall for each station, *latitude* and *longitude* for each station, distance $d$, components $n$ in PCA
**Output**: clusters and their stations
1: ***scaled_rain*** ← Calculate the average monthly rainfall for each station as training data and normalise the data
2: ***data_for_clustering*** ← Reduce dimensionality using PCA with $n$ components given *latitude, longitude* and *scaled_rain*
3: ***clusters*** ← Calculate the Euclidean distance and determine clusters from *data_for_clustering* where the Euclidean distance between cluster centroid < distance $d$

## III. OPTIMISED NORTH-EAST MONSOON AREAS USING REINFORCEMENT LEARNING

Reinforcement learning (RL) is a class of machine learning model that can optimise an objective function by maximising reward values through agent interactions. It has been proven effective in optimising complex problems compared with rule-based methods. Deep Q-Network (DQN) [9] and Deep Deterministic Policy Gradient (DDPG) [10] are the most commonly used RL models. DDPG is designed for continuous action spaces, whereas DQN is suitable for discrete action spaces. Therefore, we choose DQN in this work to optimise the geographic SST areas used to calculate the NEM index.

### *A. Deep Q-Networks*

DQNs formalise sequential decision-making problems in which an agent interacts with an environment by selecting actions to maximise the expected cumulative reward. Value-based RL methods learn an action–value function $Q(s, a)$, which estimates the expected return from taking action $a$ in state $s$ while following the policy thereafter. The learning update of $Q(s, a)$ follows temporal-difference learning and converges under tabular representations, given an exploration probability.

In our setting, we employ a DQN with a discount factor of 0.99, a final exploration probability = 0.1 and timesteps = 100,000 to approximate the function $Q(s, a)$. The action space is discrete, consisting of incremental shifts or resizing operations of SST rectangles.

### *B. Season-aware objective functions*

From subsection 2, we formulate an objective function to optimise the areas of the NEM used to calculate $Q(s, a)$. A skillful index should capture the correlations between rainfall during the NEM onset in southern Thailand and rainfall during the NEM retreat in other areas of Thailand. Accordingly, we optimise a season-aware objective function based on the squared correlation ($R^2$) between a learned NEM index and cluster-aggregated rainfall in Thailand.

Formally, let A and B be sets of rectangular areas. We define the average area SST over areas A and B at time t as $\mathrm{SST_A(t)}$ and $\mathrm{SST_B(t)}$, respectively. Hence, the candidate NEM index is calculated as:

$$Z(t) = normalise\,(\mathrm{SST_B(t)} - \mathrm{SST_A(t)}) \quad (1)$$

Consequently, we propose the season-aware objective function to be optimised as:

$$Q(A, B) = \frac{1}{2}(R^2_{NEM\ onset} + R^2_{NEM\ retreat}) \quad (2)$$

$$R_{NEM\ onset} = corr(Z(t), Y_{NEM\ onset}\,(t));\ \mathrm{t} \in NEM\ onset \quad (3)$$

$$R_{NEM\ retreat} = corr(Z(t), Y_{NEM\ retreat}(t));\ \mathrm{t} \in NEM\ retreat \quad (4)$$

where $Y_{NEM\ onset}(t)$ denotes the average rainfall across stations in southern Thailand, which is influenced by the onset of the NEM, $NEM\ onset = \{\mathrm{October} - \mathrm{March}\}$, and $Y_{NEM\ retreat}(t)$ denotes average rainfall across stations in upper Thailand, which is influenced by the retreat of the NEM, $NEM\ retreat = \{\mathrm{April} - \mathrm{September}\}$. The ocean-coverage constraints for the rectangular areas A and B are:

$$min\ ocean_fraction(A), ocean_fraction(B) \geq 0.8$$

### *C. Datasets and environment settings*

We define two agent action configurations processed from Optimum Interpolation Sea Surface Temperature (OISST) [11] for the DQN environment: (1) shift-only and (2) shift-and-resize. In the shift-only configuration, the agent shifts areas A and B by ±0.5° in latitude and ±0.5° in longitude. This configuration results in a total of eight discrete actions. Alternatively, we employ a shift-and-resize configuration, in which areas A and B can be both shifted and resized. In this case, the action space increases to 16 actions, including shrinking and expanding the boundaries of areas A and B by ±0.5° in latitude and ±0.5° in longitude. A geometric constraint is imposed to ensure that the spatial extent of areas A and B remains greater than zero.

We use monthly rainfall data measured at gauge stations from two agencies: 74 high-quality Thai Meteorological Department (TMD) gauge stations from 1982-2024 and 384 high-quality Hydro-Informatics Institute (HII) stations from 2014-2024. We exclude any stations that report less than 80% of their data for each month across the study period. Missing rainfall values for each station are imputed using the monthly median. All HII stations are used for clustering in Algorithm 1 as they cover most areas in Thailand, whereas TMD stations are used to train the DQN RL agent as the dataset spans 43 years. Finally, HII stations are used to report the average Pearson correlation performances as test datasets over 11 years for each period of the NEM onset and retreat.

To evaluate the predictive skill of long-term rainfall forecasting when incorporating the optimised NEM index, we use the Long Short-Term Memory (LSTM) model [12] with a 24-month input window and a 12-month prediction horizon based on TMD rainfall observations. The training dataset consists of two folds. Fold 1 utilized training data from 1982–2019, with 2020 as the validation year and 2021 as the test year while Fold 2 utilized training data from 1982–2022, with 2023 as the validation year and 2024 as the test year. Model performance is assessed using the Root Mean Squared Error (RMSE). RMSE is defined as $RMSE = \sqrt[2]{MSE}$ and $MSE = \frac{\sum (y_{i,t} - \hat{y}_{i,t})^2}{n}$ where $y_{i,t}$ denotes the observed rainfall at station $i$ and time $t$, $\hat{y}_{i,t}$ is the predicted rainfall, and $n$ represents the total number of stations and time steps. The training hyperparameters are as follows: hidden layer size = [16,32,64]; number of epochs = 200 with early stopping on the validation set; number of LSTM layers = [1, 2, 3]; dropout rates = [0, 0.2, 0.5]; and a learning rate of 0.01. All experiments are implemented in PyTorch using Python 3.8 on a high-performance computing (HPC) system equipped with a 32-core Intel Xeon CPU (2.3 GHz), 64 GB of RAM, and NVIDIA V100 GPUs.

## IV. EXPERIMENT RESULTS

### A. *Rainfall Clusters in Thailand*

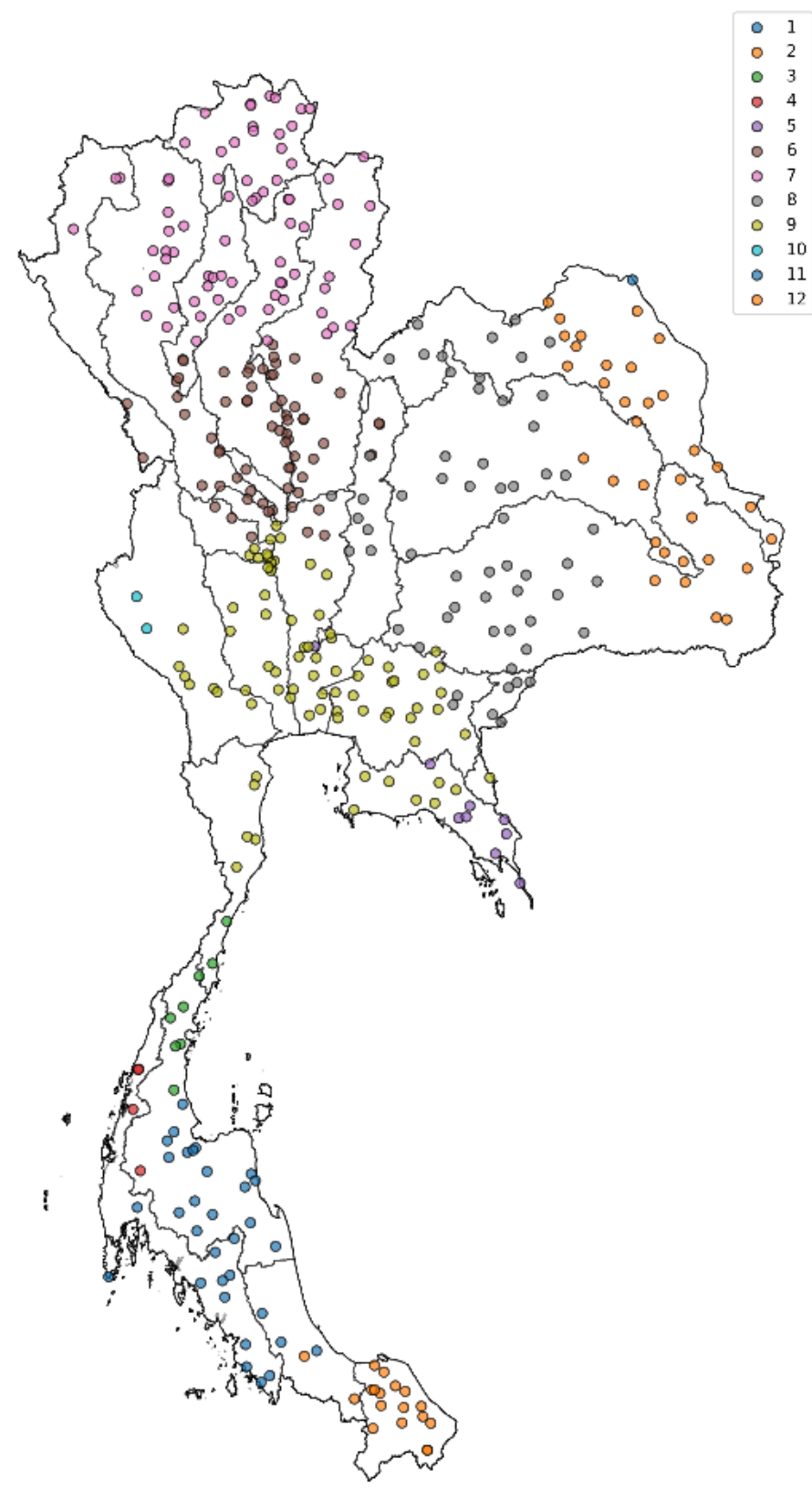

Fig. 1. Twelve clusters of HII rainfall stations obtained by Algorithm 1

Fig. 1 shows 12 clusters of 384 HII stations from the hierarchical clustering algorithm in Algorithm 1 with $d = 2$, $n = 2$. The results illustrate that Clusters 1-4 refer to station areas in southern Thailand, while Clusters 5-12 represent areas in upper Thailand including northern, northeastern, western, eastern, and central Thailand. We then applied these clustering areas to TMD stations for optimising NE areas by calculating the objective function in DQN.

### B. *Optimised areas with the objective score*

In Table 1, the DQN agent was trained to optimise the SST calculation areas (Areas A and B) across two settings: shift-only and shift-and-resize. The shift-only setting achieved the highest objective score (Q=0.497), showing a significant improvement over the initial baseline areas from the previous study [4] (Q=0.052), as shown in Fig. 2. Meanwhile, the shift-and-resize setting yielded the objective score of 0.412, as shown in Fig. 3. Table 2 illustrates that the optimised NEM index with the shift-only setting provides high correlations during both the NE onset and retreat for each cluster of test rainfall data from HII stations. Specifically, in the NE onset, strong negative correlations were observed in southern Thailand, particularly in Cluster 1 (R=−0.720) and Cluster 4 (R=−0.645). During the NEM retreat, the index maintained high correlations with rainfall across upper Thailand, such as Cluster 5 (R=−0.738) and Cluster 8 (R=−0.632). Additionally, dividing the study area by 101°E longitude revealed that the NEM index strongly influences the left side of Thailand during the onset period (R=−0.718) and the right side of Thailand during the retreat period (R=−0.711).

TABLE I. OPTIMISED AREAS WITH SETTINGS FOR TMD STATIONS

| Setting | $Q$ | $R$ (TMD stations) | | Optimised Areas | |
|---|---|---|---|---|---|
| | | NEM onset | NEM retreat | A | B |
| Initial areas [2] | 0.052 | 0.270 | -0.177 | [10,16.25,110,118.75]<br>[8.75,10,108.75,116.25]<br>[7.5,8.75,107.5,115] | [3.75,6.25,103.75,106.25] |
| Shift-and-resize | 0.412 | -0.560 | -0.714 | [18.5, 23.75,115.0,118.75]<br>[17.25,17.5,113.75,116.25]<br>[16.0,16.25,112.5,115.0] | [2.75,12.25,108.25,109.75] |
| Shift-only | **0.497** | **-0.653** | **-0.754** | [15.0,21.25,111.0,119.75]<br>[13.75, 15.0,109.75,117.25]<br>[12.5, 13.75,108.5,116.0] | [9.75,12.25,100.25,102.75] |

TABLE II. CORRELATIONS BETWEEN THE OPTIMISED NEM INDEX AND RAINFALL FROM HII STATIONS

| Cluster | $R$ HII stations (test data) | |
|---|---|---|
| | NE onset (Oct–Mar) | NE retreat (Apr–Sep) |
| 1 | **-0.720** | -0.386 |
| 2 | **-0.607** | -0.272 |
| 3 | -0.548 | -0.414 |
| 4 | **-0.645** | **-0.766** |
| 5 | -0.192 | **-0.738** |
| 6 | -0.315 | **-0.604** |
| 7 | -0.340 | -0.551 |
| 8 | -0.267 | **-0.632** |
| 9 | -0.545 | **-0.618** |
| 10 | -0.385 | -0.545 |
| 11 | 0.274 | -0.472 |
| 12 | -0.131 | **-0.617** |
| Left of Thailand (101°E) | **-0.718** | -0.670 |
| Right of Thailand (101°E) | -0.337 | **-0.711** |

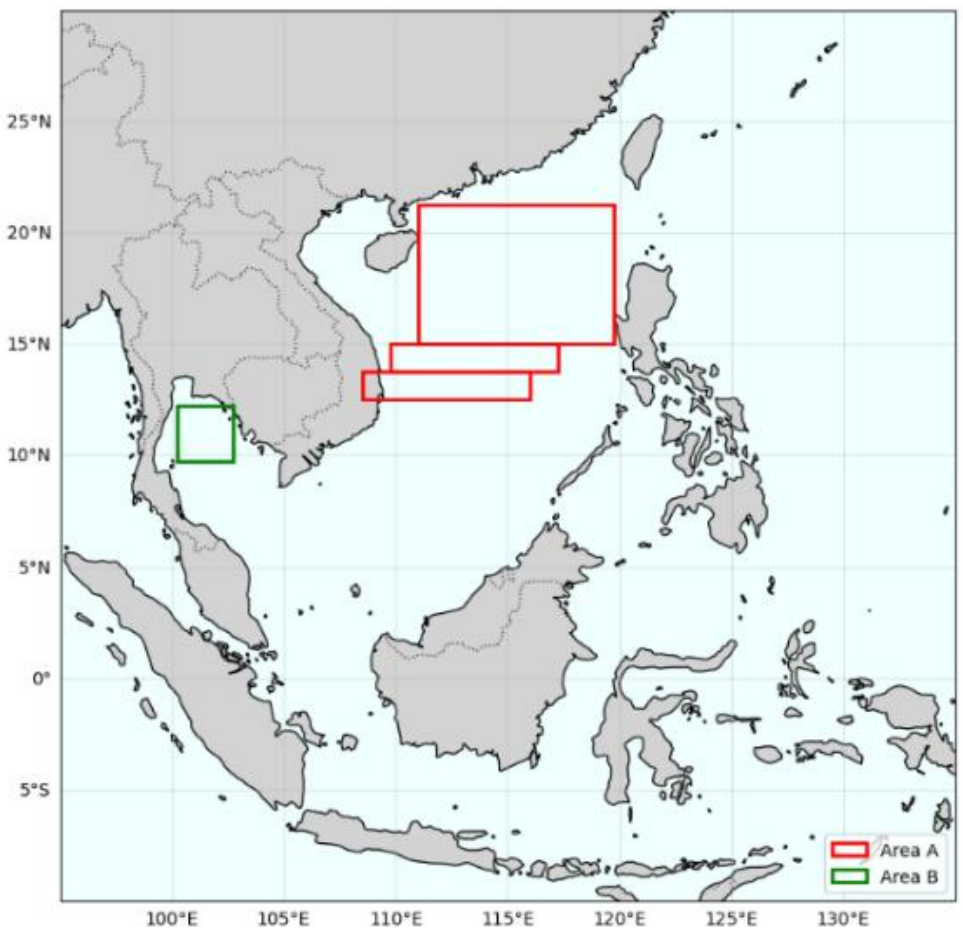

Fig. 2. Optimised NEM
areas using DQN with the shift-only setting

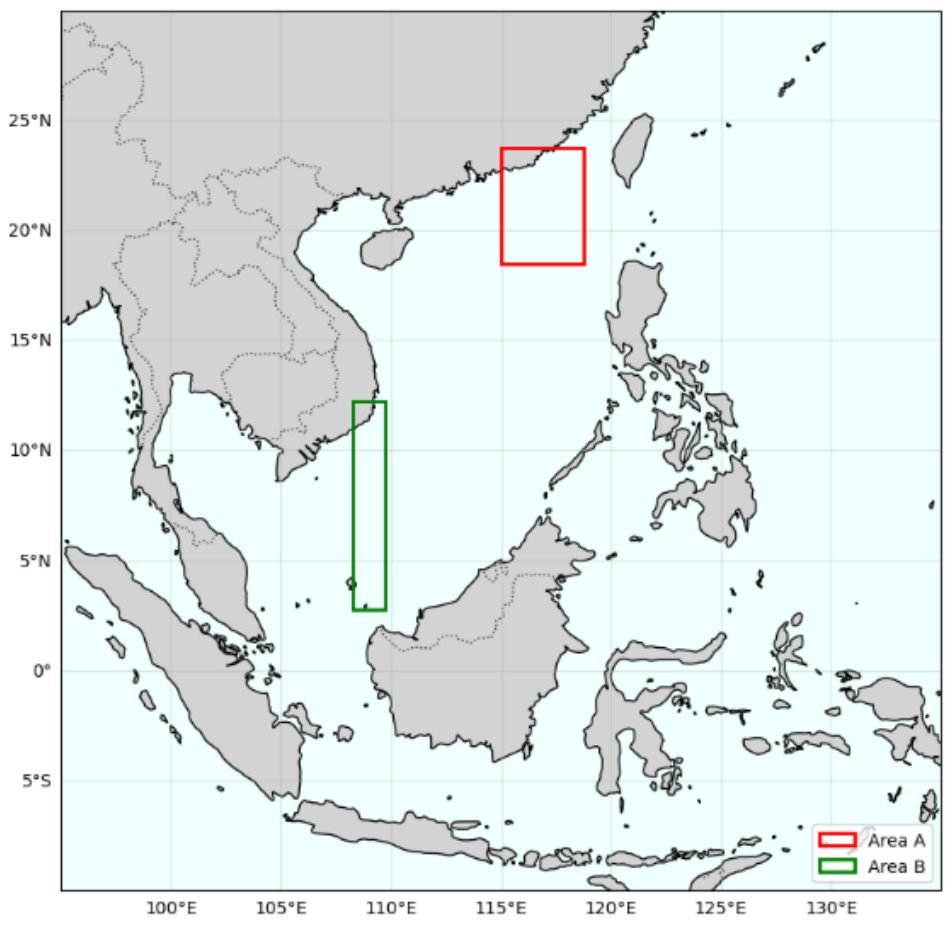

Fig. 3. Optimised NEM
areas using DQN with the shift-and-resize setting

### *C. Monthly Rainfall Predictive Performance of the Optimised NEM Index*

To evaluate long-term rainfall prediction in southern and upper Thailand, we assessed the performance for Clusters 1–12 using the features described in subsections 1 and 3. Input features were selected individually based on an absolute correlation exceeding 0.6 with the average rainfall of each cluster. The selected features include DMI, MEI, PDO, MJO, BSISO, South-West monsoon index, and ONI. Average performance metrics were reported over two test folds, comparing LSTM results with and without the incorporation of the proposed NEM index. Table 3 demonstrates that incorporating the optimised NEM index improved the RMSE for 12-month-ahead rainfall predictions. Specifically, in southern Thailand, the predictive skill improved for Clusters 1, 2, and 4, where the RMSE improved from 99.79, 82.61, and 130.02 to 94.54, 77.05, and 121.48, respectively. The NEM index also consistently improves long-term rainfall predictions in other regions of Thailand. For example, the RMSE improved from 57.27, 56.11, and 59.24 to 53.94, 55.44, and 52.38 in Clusters 6, 9, and 12, respectively. Note that the results of Clusters 3, 7, 10, and 11 are not included in the experiment because the absolute correlations of the NEM index are lower than 0.6.

TABLE III. PREDICTIVE SKILLS OF RAINFALL PREDICTION IN THAILAND USING LSTM WITH AND WITHOUT THE NEM INDEX

| Cluster | RMSE of LSTM (mm/month) | |
|---|---|---|
| | Selected features | Selected features + NEM |
| 1 | 99.79 | **94.54** |
| 2 | 82.61 | **77.05** |
| 4 | 130.02 | **121.48** |
| 5 | 145.12 | **136.84** |
| 6 | 57.27 | **53.94** |
| 8 | **59.81** | 59.91 |
| 9 | 56.11 | **55.44** |
| 12 | 59.24 | **52.38** |

## V. CONCLUSION

This work proposes a novel NEM climate index using DQN reinforcement learning agents to optimise SST-based calculation areas. The study formulates a season-aware objective function that captures rainfall relationships during both the NEM onset (October–March) and the NEM retreat (April–September) within specific impact regions. The findings indicate that the index provides superior predictive skill compared to models that rely solely on traditional climate indices when using LSTM models. These results suggest that identifying local-scale patterns is crucial for effective long-term rainfall prediction for water management. Future work should employ spatiotemporal models, such as Convolution-LSTM, Graph Neural Networks, Neural Operators, or state-space models to predict long-term rainfall using the index.